\documentclass{article}
\PassOptionsToPackage{numbers,sort&compress}{natbib}
\usepackage[preprint]{neurips_2026}

\usepackage[utf8]{inputenc}
\usepackage[T1]{fontenc}
\usepackage{amsmath,amssymb,amsfonts}
\usepackage{booktabs}
\usepackage{graphicx}
\usepackage{microtype}
\usepackage{placeins}
\usepackage{float}
\usepackage{url}
\usepackage{hyperref}
\graphicspath{{figures/}}
\hypersetup{hidelinks}

\title{Do Better Scores Mean Better Physics? Physics-Grounded Explanations for Sim2Real Neural Operators%
\thanks{Accepted at the NeurIPS 2026 XAI4Science Workshop (Tiny Paper Track). This is a non-archival workshop preprint, not a NeurIPS main-conference paper.}}

\author{
Somyajit Chakraborty \quad Xizhong Chen\\
State Key Laboratory of Synergistic Chem-Bio Synthesis, \\
Department of Chemical Engineering, School of Chemistry and Chemical Engineering, \\
Shanghai Jiao Tong University, Shanghai 200240, People's Republic of China \\
\texttt{chksomyajit@sjtu.edu.cn},\quad
\texttt{chenxizh@sjtu.edu.cn}
}

\newcommand{\rel}{\mathrm{Rel}\text{-}L_2}
\newcommand{\tke}{\mathrm{TKE}}

\begin{document}
\maketitle

\begin{abstract}
Machine-learning surrogates accelerate physical simulation, but lower prediction error need not coincide with lower error in physically relevant flow statistics. We examine this question for flow around a NACA4418 airfoil using paired computational-fluid-dynamics simulations and experimental particle-image-velocimetry measurements. A mean-preserving input intervention removes velocity fluctuations from selected regions of observed flow histories. Across four neural operators, removing fluctuations from the most energetic 10\% of valid observed cells changes forecasts more than equal-area random removal. Because the masks are not matched for removed fluctuation energy, this contrast measures sensitivity, not independent evidence of physical importance. Separately, a CNO has lower velocity-field error but substantially higher two-component fluctuation-energy error than the reference on both analysis subsets. An output attenuation stress test also demonstrates disagreement between benchmark errors and domain-summed fluctuation energy. These single-benchmark results motivate reporting complementary physical diagnostics alongside aggregate prediction scores; they do not establish counterfactual physical correctness.
\end{abstract}

\section{Introduction}
Neural operators learn mappings between fields and are increasingly used as fast surrogates for parameterized PDEs and time-dependent physical systems \citep{li2021fno,lu2021deeponet,kovachki2023neuraloperator}. In fluid mechanics, however, a small field error does not guarantee accurate fluctuation dynamics or coherent structures. Therefore, turbulence-oriented operator studies increasingly report flow statistics beyond pointwise errors \citep{li2023iufno,wang2024channel}. This motivates diagnostics that ask \emph{what} physically changes when a score improves.

The issue is especially important in Sim2Real transfer, where a surrogate is trained or adapted using simulation but ultimately judged against measurements. RealPDEBench provides broader paired simulation--measurement context \citep{hu2026realpdebench}, while the NeurIPS 2026 RealPDE Competition releases a NACA4418 paired PIV/CFD task for Sim2Real forecasting \citep{wu2026neurips}. In parallel, perturbation-based XAI makes feature importance testable by altering selected inputs and measuring the model response, with matched random removal providing a useful faithfulness control \citep{fong2017meaningful,hooker2019benchmark}. Beyond predictive accuracy, scientific machine learning requires evaluating whether models preserve meaningful physical relationships. A physically trustworthy surrogate should not only reproduce future states, but should respond consistently when physically meaningful components of the input are modified. This motivates physically defined diagnostic interventions, while recognizing that model sensitivity to an intervention is not itself evidence of physically correct behavior.

We ask two connected questions: \emph{how do forecasts respond to interventions in energetic input regions, and can improved field accuracy coexist with worse fluctuation-energy diagnostics?} First, we compare mean-preserving removal of temporal velocity fluctuations in high-energy regions with equal-area random and low-energy controls. Second, we quantify this response across mask fractions and four tested architectures: a hybrid reference, FNO \citep{li2021fno}, CNO \citep{raonic2023cno}, and Transolver \citep{wu2024transolver}. Third, we document disagreements between velocity-field and fluctuation-energy metrics. The goal is a diagnostic protocol, not a new architecture or a demonstration of causal physical importance.

\begin{figure}[t]
  \centering
  \includegraphics[width=\linewidth]{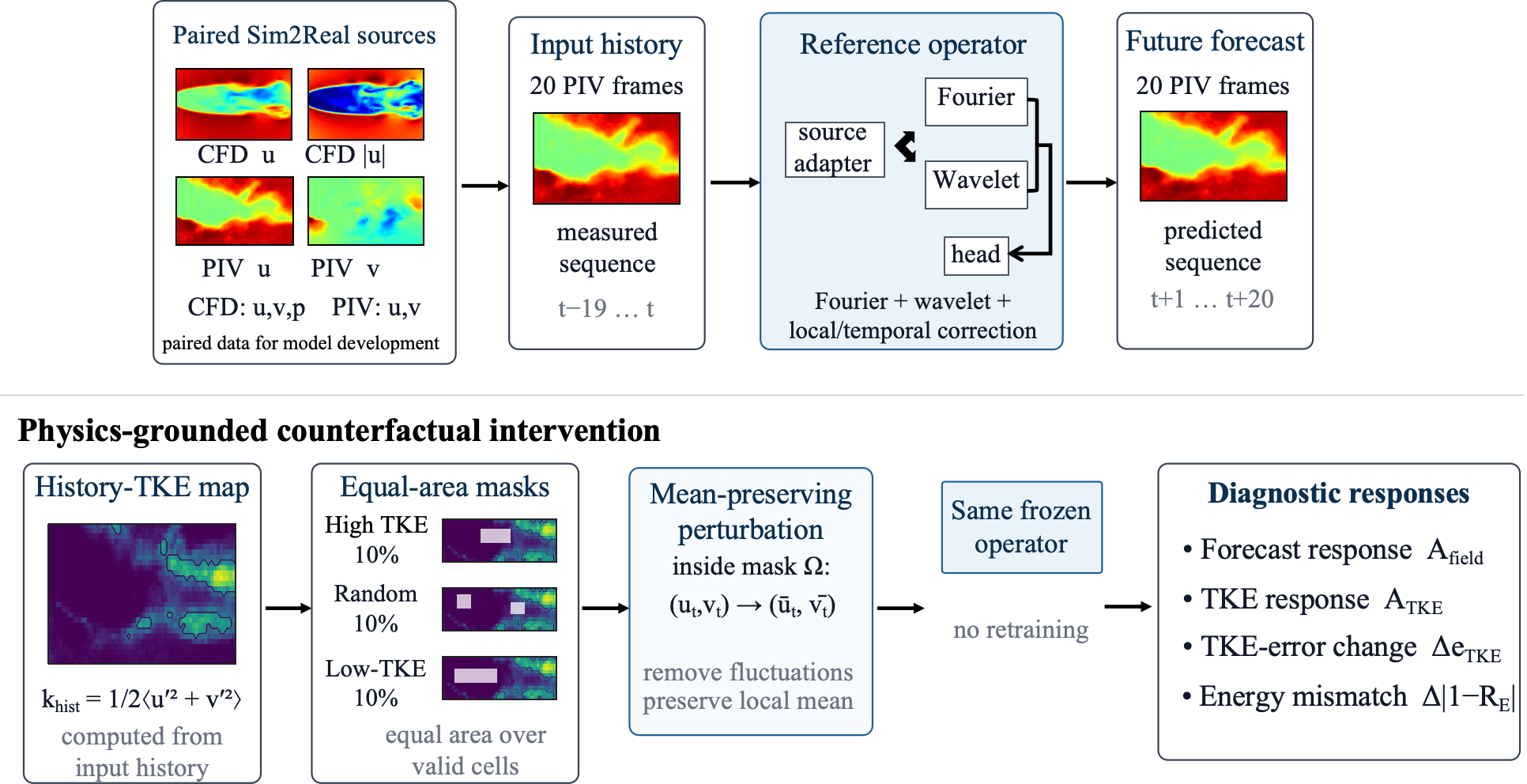}
  \caption{Sim2Real forecasting and physics-grounded explanation workflow. Paired CFD/PIV data support model development, while real inference maps 20 observed PIV frames to 20 future frames. History fluctuation energy defines equal-area energetic, random, and low-energy masks; masked fluctuations are replaced by their local temporal mean, and the frozen surrogate response is evaluated. Equal area does not imply equal removed fluctuation energy.}
  \label{fig:workflow}
\end{figure}

\section{Data, models, and diagnostic protocol}
\paragraph{Dataset and models.}
We use the Sim-to-Real Transfer Learning task of the NeurIPS 2026 RealPDE Competition \citep{wu2026neurips}. The release contains paired time-resolved PIV measurements and matched CFD simulations around a NACA4418 airfoil, with angles of attack $0^\circ$--$20^\circ$, Reynolds numbers 2968--27975, 100 paired trajectories, roughly 600 time steps per trajectory, and $64\times128$ released fields. CFD provides $u,v,p$; PIV provides measured $u,v$. The scorer used here retains velocity after $2\times$ spatial subsampling, giving $32\times64$. Each sample maps 20 history frames to 20 future frames. We analyze 184 windows (historically labelled ``selection'') and another 768 windows (``calibration''). These labels denote analysis bookkeeping, not independent model selection and validation: windows do not overlap, but both subsets draw from the same 23 experimental trajectories and operating-condition support. They provide within-trajectory replication rather than out-of-trajectory generalization. Our conclusions concern those 23 trajectories, not all 100 released trajectories. Statistical intervals use 5,000 percentile bootstrap replicates over complete trajectory IDs.

The reference is a compact hybrid Fourier--wavelet operator with local convolution, temporal attention, source adaptation, and residual forecasting components; related Fourier--wavelet CFD-surrogate design is described in \citep{chakraborty2026pibert}. Here ``reference'' means the chosen comparison checkpoint, not a physics-ground-truth model. To test architecture dependence, we repeat the counterfactual on released real-finetuned FNO, CNO, and Transolver baselines under identical windows, masks, and controls. Their prescribed normalization and output post-processing are retained; inference-cache parity on measured $u,v$ is documented in the appendix.

\paragraph{Benchmark errors.}
For sample $n$, let $\mathbf U_n$ denote the measured future in-plane velocity tensor over time, space, and channels $(u,v)$. Following competition terminology,
\begin{equation}
\begin{aligned}
 e_{\rel}&=\frac1N\sum_n\frac{\|\hat{\mathbf U}_n-\mathbf U_n\|_2}{\|\mathbf U_n\|_2},\qquad
 k_n(x,y)&=\tfrac12\langle(u-\bar u)^2+(v-\bar v)^2\rangle_t,\\
 e_{\tke}&=\frac1N\sum_n\frac{\|\hat k_n-k_n\|_2}{\|k_n\|_2}.
\end{aligned}
\label{eq:benchmark}
\end{equation}
Here particle-image velocimetry (PIV) measures the two in-plane velocity components; ``TKE'' denotes their resolved temporal fluctuation energy, \emph{not} full three-component turbulent kinetic energy. The competition scorer evaluates velocity-field and fluctuation-energy errors; its Mean Velocity Profile Error (MVPE) is a relative error of time-averaged velocity profiles at prescribed probe locations \citep{wu2026neurips}.

\paragraph{Physics-grounded counterfactual attribution.}
For each observed 20-frame PIV history we compute $k_{\mathrm{hist}}(x,y)=\tfrac12\langle(u-\bar u)^2+(v-\bar v)^2\rangle_{\mathrm{hist}}$. Among valid observed fluid cells, equal-area masks select the highest-energy 10\%, lowest-energy 10\%, or a matched random 10\%; ten deterministic random masks are evaluated per window. Inside mask $\Omega$ we replace $(u_t,v_t)\rightarrow(\bar u,\bar v)$, removing temporal fluctuation information while preserving the local temporal mean. The same frozen surrogate is rerun without retraining. Model response is
\begin{equation}
 A_{\mathrm{field}}=\frac{\|\hat U_{\Omega}-\hat U_0\|_2}{\|\hat U_0\|_2},\qquad
 A_{\mathrm{TKE}}=\frac{\|\hat k_{\Omega}-\hat k_0\|_2}{\|\hat k_0\|_2},
 \label{eq:attribution}
\end{equation}
where subscript $0$ denotes the unperturbed forecast. Contrasts $A_H-A_R$ compare energetic-region removal with equal-area random controls. Equal mask area does \emph{not} match the fluctuation energy removed, so larger responses may reflect larger input perturbations instead of uniquely greater physical importance. For the reference we also test 5\% and 20\% masks. No physically realizable future target is available for these modified histories.

\paragraph{Complementary physical diagnostics and interventions.}
We compute the domain-summed fluctuation-energy ratio
\begin{equation}
 R_E=\frac{\sum_{n,x,y}\hat k_n(x,y)}{\sum_{n,x,y}k_n(x,y)},
 \label{eq:energy}
\end{equation}
to quantify total predicted-versus-measured resolved fluctuation energy. Mean per-sample spatial TKE correlation tests pattern alignment; pooled unnormalised squared TKE error measures absolute discrepancies; and TKE MSE inside the measured top-energy decile $\Omega_{90}$ tests errors in energetic regions. A horizon-conditioned decoder variant supplies a within-lineage comparison. Separately, an output-side stress test (a post-processing intervention, not retraining)
\begin{equation}
\hat{\mathbf u}'_t=\bar{\hat{\mathbf u}}+\mathrm{diag}(0.975,0.935)(\hat{\mathbf u}_t-\bar{\hat{\mathbf u}})
\label{eq:attenuation}
\end{equation}
preserves the temporal mean exactly while reducing predicted fluctuation amplitudes. Its gains were selected using both analysis subsets, so attenuation is interpreted only as a diagnostic stress test, not a held-out correction.

\section{Results}
\subsection{Energetic history fluctuations consistently influence forecasts across models}
\begin{figure}[t]
  \centering
  \includegraphics[width=0.85\linewidth]{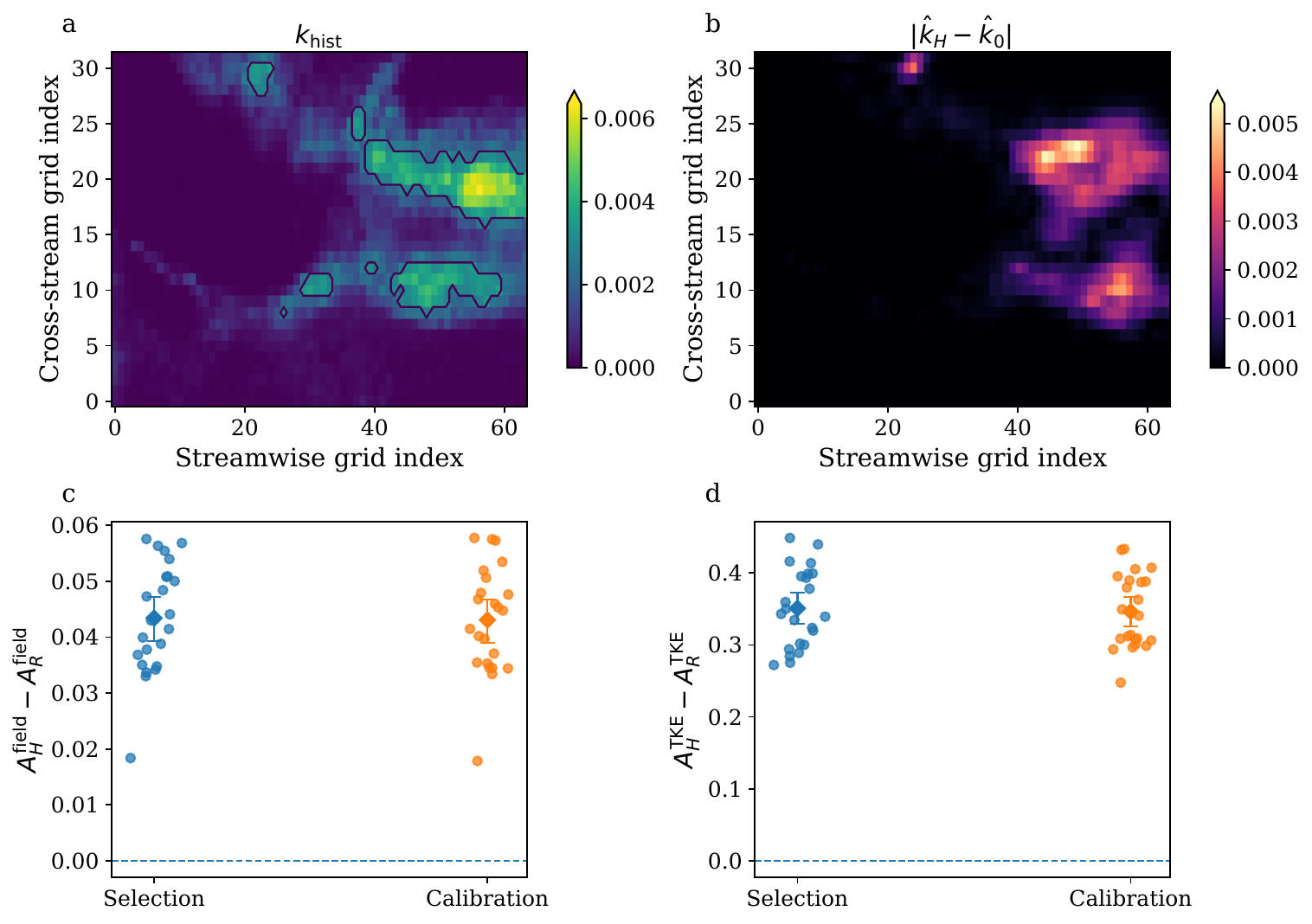}
  \caption{Physics-grounded counterfactual attribution for the reference surrogate. (a) History fluctuation-energy map $k_{\mathrm{hist}}$ for a representative calibration window; the contour marks the highest-energy 10\% of valid observed cells. (b) Absolute change in predicted future TKE after removing temporal fluctuations inside that region. (c,d) Trajectory-level energetic-minus-random contrasts for future field response and TKE-map response. Points are equal-weight trajectory aggregates; diamonds and bars denote the mean and 95\% trajectory-bootstrap interval.}
  \label{fig:xai}
\end{figure}

For the reference surrogate, energetic-region removal changes the future velocity forecast more than matched random masking by $A_{\mathrm{field}}^{H}-A_{\mathrm{field}}^{R}=0.0434$ (95\% trajectory-bootstrap CI $[0.0393,0.0471]$) on selection and $0.0431$ ($[0.0390,0.0468]$) on calibration. The corresponding TKE-response contrast is $0.3508$ ($[0.3294,0.3726]$) and $0.3463$ ($[0.3259,0.3663]$), with all 23 trajectory aggregates positive on both subsets. Relative to random masking, energetic removal also increases target-evaluated TKE error by $0.1332$ ($[0.1099,0.1555]$) and $0.1326$ ($[0.1123,0.1513]$), with 22 of 23 trajectories positive (Fig.~\ref{fig:xai}).

The finding is not specific to one threshold or one architecture. For the reference, $A_{\mathrm{TKE}}^{H}-A_{\mathrm{TKE}}^{R}$ remains positive at 5\%, 10\%, and 20\% masks, with all 23 trajectory aggregates positive on both subsets. At the canonical 10\% mask, FNO, CNO, and Transolver independently reproduce the energetic-over-random TKE response; all 23 trajectory aggregates are again positive on each subset. The selection/calibration contrasts are respectively 0.423/0.421 for FNO, 0.397/0.392 for CNO, and 0.493/0.482 for Transolver (Appendix~\ref{app:robustness}). Thus the response contrast repeats across these four surrogate families, although the equal-area test cannot distinguish spatial importance from removed-energy magnitude.

\subsection{Field error and TKE error can rank models oppositely}
\begin{figure}[t]
  \centering
  \includegraphics[width=0.80\linewidth]{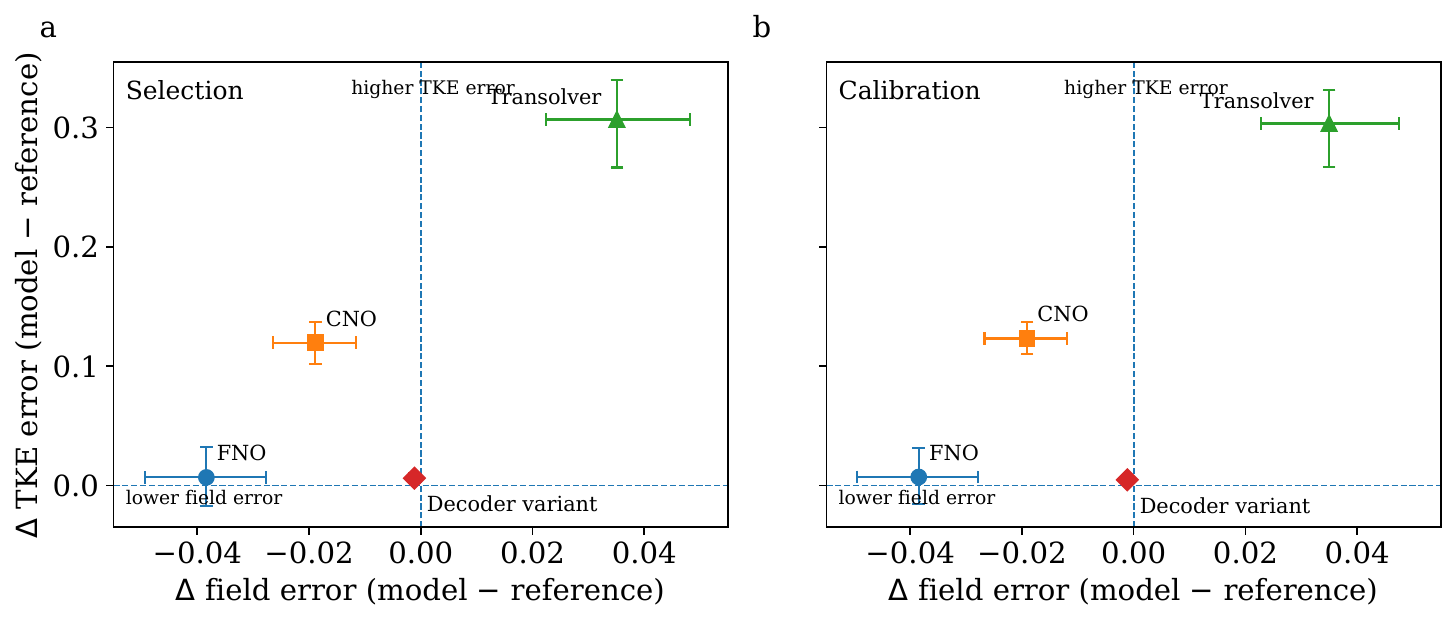}
  \caption{Cross-model metric ranking. Points are model-minus-reference trajectory-balanced means with 95\% bootstrap intervals. The upper-left quadrant denotes lower field error but higher TKE error: CNO lies robustly there on both subsets, while the decoder variant gives a smaller within-lineage example.}
  \label{fig:conflict}
\end{figure}

The disagreement between field accuracy and fluctuation-energy accuracy is not restricted to a small modification of the reference. A separately trained CNO achieves lower field error but substantially higher TKE error on both subsets. On selection, the trajectory-balanced model-minus-reference differences are $\Delta e_{\rel}=-0.0189$ (95\% CI $[-0.0265,-0.0117]$) and $\Delta e_{\tke}=+0.1196$ ($[0.1016,0.1369]$); calibration gives $-0.0191$ ($[-0.0267,-0.0120]$) and $+0.1232$ ($[0.1101,0.1366]$). The opposite ranking occurs for 18 of 23 trajectory aggregates on each subset. FNO lowers field error even more strongly, but its small positive TKE-error difference is not resolved at trajectory level; Transolver is worse in both metrics (Fig.~\ref{fig:conflict}).

The controlled decoder variant supplies a smaller within-lineage replication: $\Delta e_{\rel}<0$ while $\Delta e_{\tke}>0$, with trajectory-bootstrap intervals excluding zero in opposite directions on both subsets. Together, the cross-architecture and within-lineage comparisons show that lower pointwise field error does not determine whether fluctuation-energy prediction has improved.

\subsection{Lower TKE error can coexist with a larger fluctuation-energy deficit}
\begin{figure}[H]
  \centering
  \includegraphics[width=0.85\linewidth]{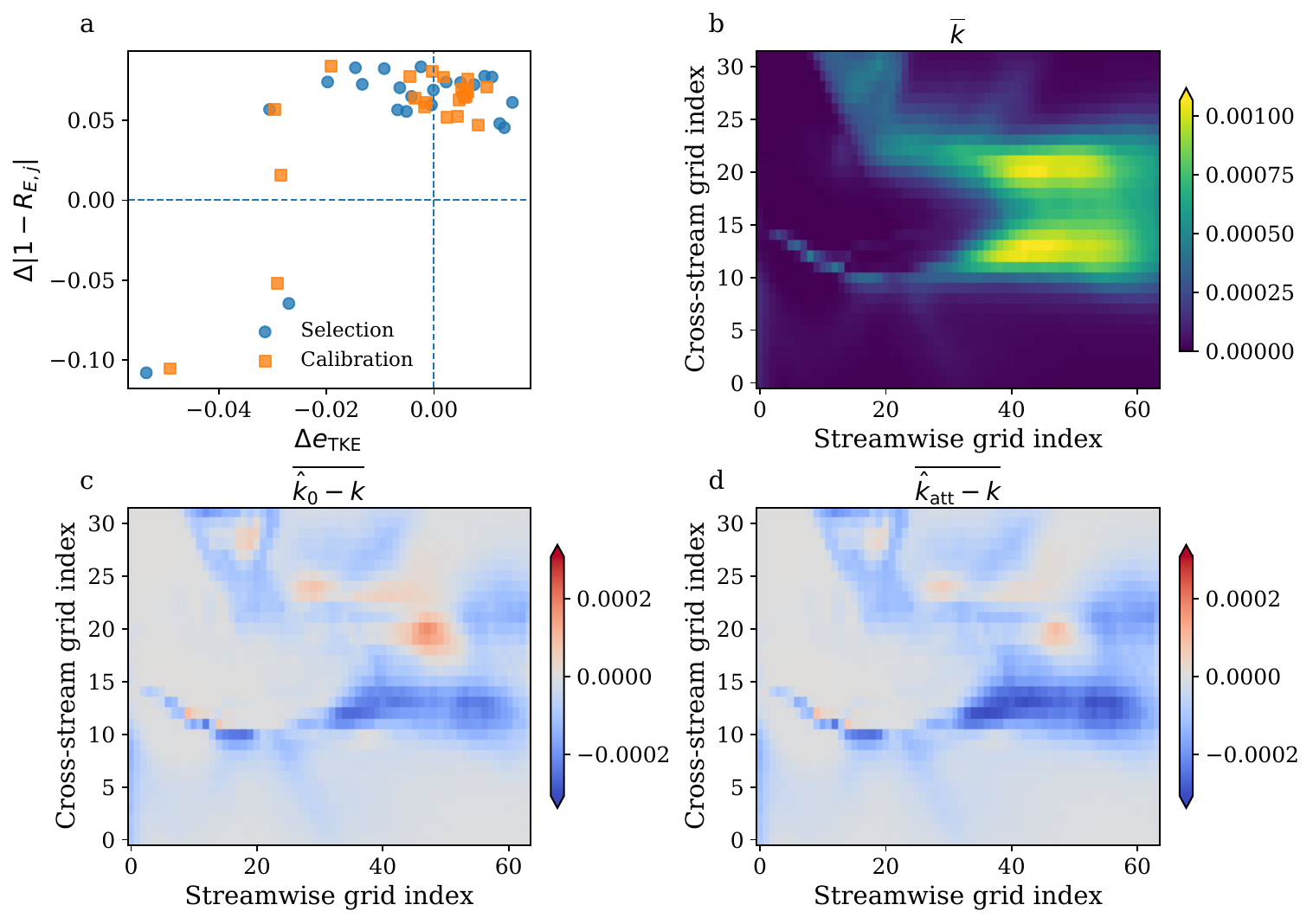}
  \caption{Benchmark TKE error and domain-summed energy mismatch rank attenuation differently. (a) Trajectory-level $\Delta e_{\mathrm{TKE}}$ versus $\Delta|1-R_{E,j}|$. (b) Calibration-mean TKE. (c,d) Signed TKE bias before and after attenuation.}
  \label{fig:attenuation}
\end{figure}

The attenuation provides a diagnostic example where lower benchmark error coexists with larger error in total resolved fluctuation energy; it does not establish a general physically correct ranking. On selection $e_{\rel}$ changes $0.120958\rightarrow0.119692$ and $e_{\tke}$ $0.533389\rightarrow0.528463$; calibration gives $0.121309\rightarrow0.120057$ and $0.533985\rightarrow0.530093$. Yet the reference is already under-energetic, with $R_E\simeq0.815$, and attenuation lowers it to $R_E\simeq0.756$. Target-top-decile TKE MSE rises by 3.9\% and 3.7\%, while MSE over the remaining 90\% falls by about 12.8\%; pooled unnormalised TKE-squared error also rises and spatial TKE correlation is essentially unchanged. Because Eq.~\eqref{eq:benchmark} averages separately normalized sample residuals whereas $R_E$ pools fluctuation energy over the domain, the two diagnostics need not rank predictions identically (Fig.~\ref{fig:attenuation}).

\section{Discussion and limitations}
These tests demonstrate intervention sensitivity and disagreement among velocity- and fluctuation-energy metrics, not counterfactual physical correctness. Equal-area masks remove unequal fluctuation energy, and hard boundaries may introduce spectral artifacts. The evidence covers one NACA4418 system and overlapping trajectory support; attenuation gains were tuned on both subsets. Appendix~\ref{app:boundaries} details these limits and future controls.

\section*{Acknowledgements}

Financial support from the National Natural Science Foundation of China
(Grant No.~22308212) and the Shanghai Municipal Education Commission
(Grant No.~2024AIYB004) is gratefully acknowledged.

\bibliography{references}
\bibliographystyle{abbrv}

\appendix
\section{Data schema and analysis subsets}
Table~\ref{tab:data_schema} summarizes the data representation and analysis subsets. The analysed data are from the NACA4418 NeurIPS 2026 RealPDE Competition release~\citep{wu2026neurips}; RealPDEBench~\citep{hu2026realpdebench} is cited only for broader paired simulation--measurement context.

\begin{table}[H]
\caption{Data and evaluation schema.}
\label{tab:data_schema}
\centering
\small
\setlength{\tabcolsep}{5pt}
\begin{tabular}{p{0.28\linewidth}p{0.64\linewidth}}
\toprule
Item & Schema / value \\
\midrule
System and domains & NACA4418 airfoil; paired CFD simulation and experimental PIV \\
Variables & CFD: $u,v,p$; PIV: measured $u,v$ \\
Operating conditions & AoA $0^\circ,5^\circ,10^\circ,15^\circ,20^\circ$; $Re=2968$--27975 \\
Released data & 100 paired trajectories; approximately 600 time steps per trajectory \\
Spatial grids & $64\times128$ released; $32\times64$ scoring grid after $2\times$ subsampling \\
Forecast window & 20 observed history frames $\rightarrow$ 20 future frames \\
Manifest fields & \texttt{path}, \texttt{start}, \texttt{length}, \texttt{domain}, \texttt{reynolds}, \texttt{aoa} \\
Analysis subsets & Selection: 184 windows; calibration: 768 windows; exact overlap: 0 \\
Shared support & 23 experimental trajectories and the same operating-condition support in both subsets \\
\bottomrule
\end{tabular}
\end{table}
The subsets are therefore window-disjoint but not trajectory- or condition-disjoint. The names ``selection'' and ``calibration'' are historical bookkeeping labels: ``calibration'' is a within-trajectory replication subset, not a previously unseen holdout. These experiments do not establish the behavior of the remaining released trajectories.

\section{Temporal forecast comparison across models}

\label{app:temporal}
\begin{figure}[t]
    \centering
    \includegraphics[width=\textwidth]{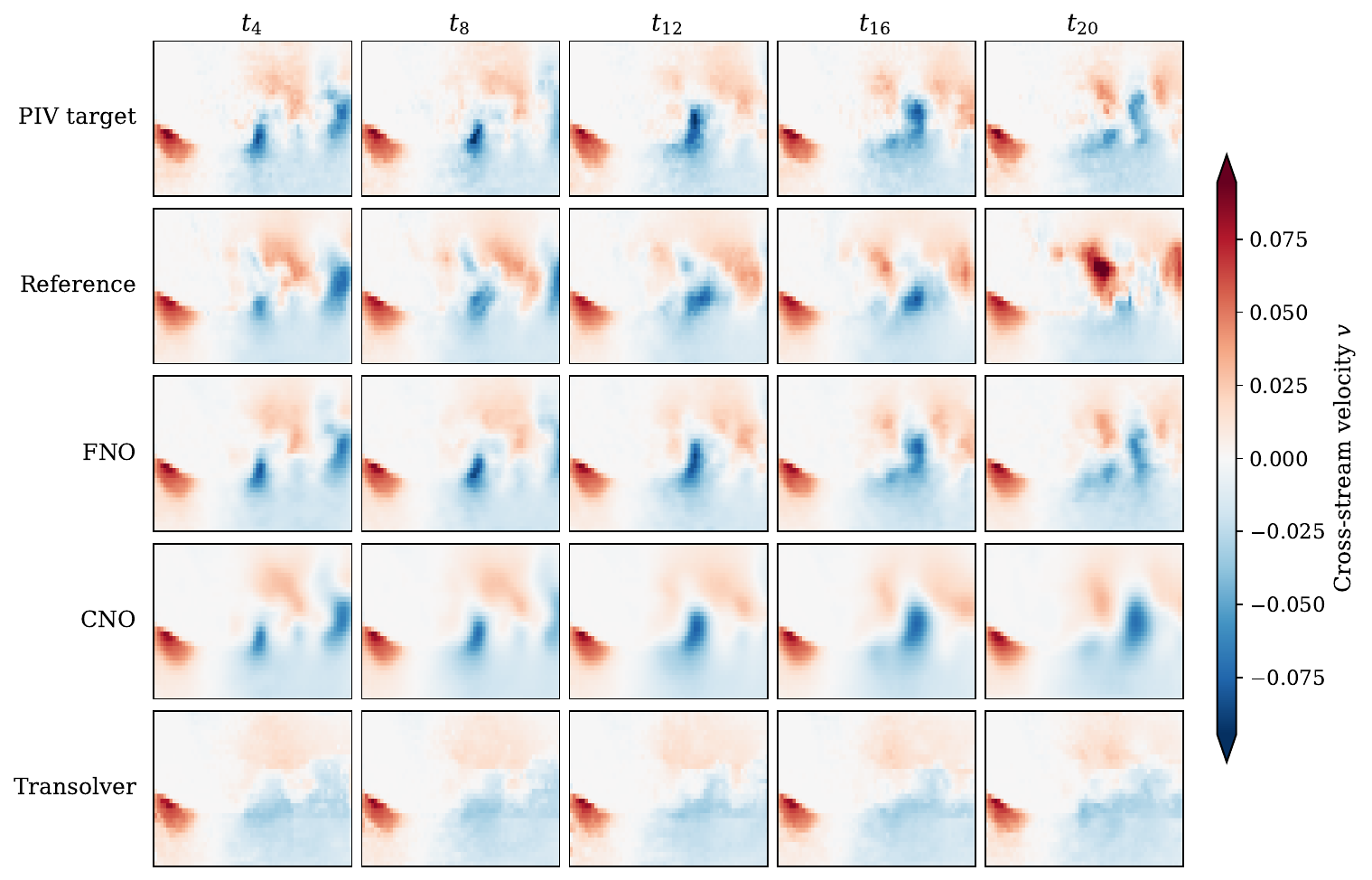}
    \caption{
    Temporal evolution of cross-stream velocity $v$ for a representative
    calibration window. The window is selected using only the reference
    surrogate as the sample whose Rel-$L_2$ is closest to the calibration
    median, and the same window is shown for every model.
    Columns correspond to future times
    $t_4,t_8,t_{12},t_{16},t_{20}$.
    All panels use the same color scale derived from the experimental PIV
    target. The comparison shows the experimental target, hybrid reference,
    FNO, CNO, and Transolver forecasts without model-specific rescaling.
    }
    \label{fig:temporal_cross_model}
\end{figure}
The representative index is selected solely from the reference model: its Rel-$L_2$ is closest to the calibration median. The same window is then reused for every model and intervention. This avoids choosing a separate visually favorable example for each architecture.

\section{Attribution robustness across mask size and model family}
\label{app:robustness}
\begin{table}[H]
\caption{Robustness of the physics-grounded attribution across perturbation size and surrogate architecture. Entries are trajectory-mean energetic-minus-random TKE-response contrasts $A_{\mathrm{TKE}}^{H}-A_{\mathrm{TKE}}^{R}$ with 95\% trajectory-bootstrap intervals.}
\label{tab:attribution_robustness}
\centering
\small
\setlength{\tabcolsep}{5pt}
\begin{tabular}{lcc}
\toprule
Setting & Selection & Calibration \\
\midrule
Reference, 5\%  & $0.285\,[0.257,0.316]$ & $0.284\,[0.259,0.312]$ \\
Reference, 10\% & $0.351\,[0.329,0.373]$ & $0.346\,[0.326,0.366]$ \\
Reference, 20\% & $0.363\,[0.347,0.378]$ & $0.358\,[0.343,0.369]$ \\
\midrule
FNO, 10\%       & $0.423\,[0.398,0.448]$ & $0.421\,[0.398,0.445]$ \\
CNO, 10\%       & $0.397\,[0.369,0.426]$ & $0.392\,[0.369,0.416]$ \\
Transolver, 10\%& $0.493\,[0.452,0.534]$ & $0.482\,[0.441,0.525]$ \\
\bottomrule
\end{tabular}
\end{table}
Every trajectory-level energetic-minus-random TKE-response contrast in Table~\ref{tab:attribution_robustness} is positive: 23/23 trajectory aggregates for every model/subset row and every reference-model mask fraction. At 10\%, the energetic-minus-random field-response contrast is also positive for all 23 trajectories in FNO, CNO, and Transolver on both subsets. Target-evaluated TKE-error changes are likewise positive for 23/23 trajectories for these three baselines. 

\begin{figure}[t]
    \centering
    \includegraphics[width=0.94\textwidth]{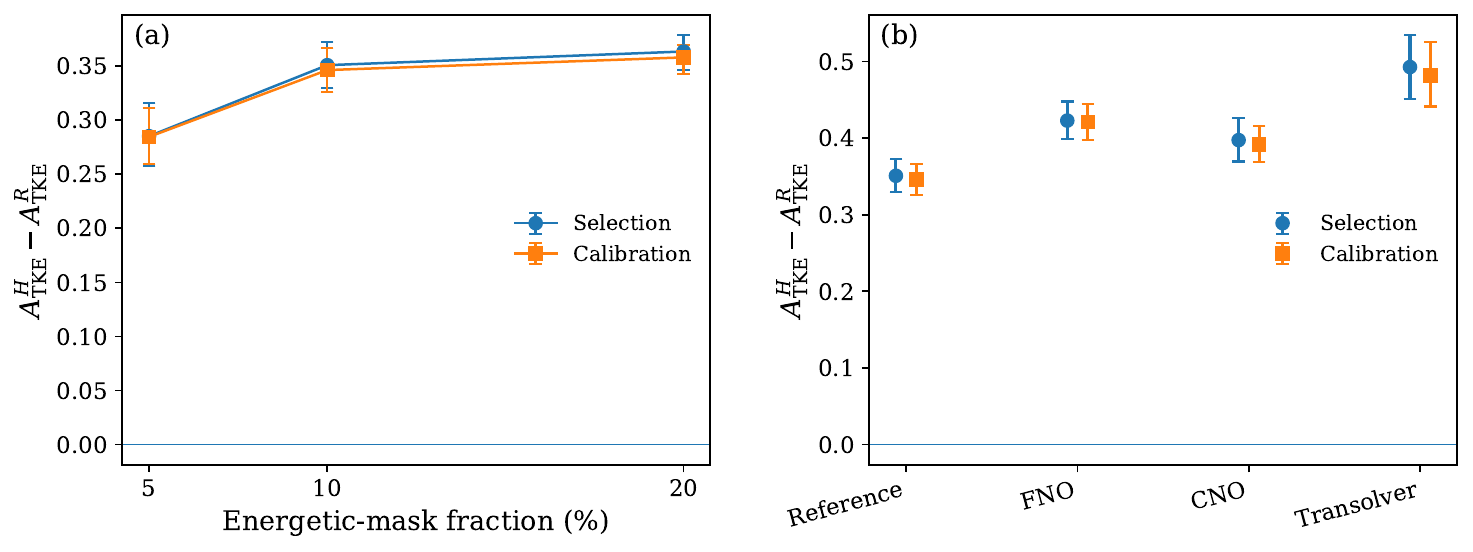}
    \caption{
    Robustness of the physics-grounded attribution.
    (a) For the hybrid reference surrogate, the energetic-minus-random
    TKE-response contrast
    $A_{\mathrm{TKE}}^{H}-A_{\mathrm{TKE}}^{R}$
    remains positive at $5\%$, $10\%$, and $20\%$ energetic-mask fractions,
    with closely agreeing selection and calibration estimates.
    (b) At the canonical $10\%$ mask fraction, the same positive contrast
    is reproduced by the reference, FNO, CNO, and Transolver.
    Error bars denote 95\% trajectory-bootstrap intervals.
    }
    \label{fig:attribution_robustness}
\end{figure}

These comparisons hold mask \emph{area}, but not removed fluctuation energy, fixed. Since the top-energy mask is constructed from the input history itself, it generally removes more fluctuation energy than a random mask. Thus cross-model agreement supports robustness of the observed \emph{response contrast} under the tested setting; it does not show that a model recognizes energetic regions beyond its sensitivity to input-perturbation magnitude. The reference-model threshold sweep additionally gives $A_{\mathrm{TKE}}^{H}-A_{\mathrm{TKE}}^{R}=0.285/0.284$ at 5\%, $0.351/0.346$ at 10\%, and $0.363/0.358$ at 20\% on selection/calibration, indicating growth followed by partial saturation rather than a sign change around the chosen 10\% operating point. Figure~\ref{fig:attribution_robustness} makes the two robustness checks
explicit. Every trajectory-level energetic-minus-random TKE-response
contrast reported in Table~\ref{tab:attribution_robustness} is positive: 23/23 trajectory aggregates for every tested model/subset combination and
for every reference-model mask fraction.

\section{Deterministic benchmark summary and cross-architecture ranking}
\begin{table}[H]
\caption{Deterministic benchmark errors. Lower values are better. FNO, CNO, and Transolver are separately evaluated released baselines; decoder and attenuation are interventions around the reference lineage.}
\label{tab:main}
\centering
\small
\setlength{\tabcolsep}{4.5pt}
\begin{tabular}{llrrr}
\toprule
Subset & Model & $e_{\rel}$ & $e_{\tke}$ & MVPE \\
\midrule
Selection & Reference & 0.120958 & 0.533389 & 0.098439 \\
& FNO & 0.082586 & 0.540132 & 0.069686 \\
& CNO & 0.102102 & 0.653011 & 0.092459 \\
& Transolver & 0.156067 & 0.840043 & 0.139438 \\
& Decoder var. & 0.119823 & 0.539500 & 0.098678 \\
& Attenuation & 0.119692 & 0.528463 & 0.098439 \\
\midrule
Calibration & Reference & 0.121309 & 0.533985 & 0.098822 \\
& FNO & 0.082562 & 0.540872 & 0.069486 \\
& CNO & 0.101990 & 0.657146 & 0.092066 \\
& Transolver & 0.156008 & 0.837650 & 0.140342 \\
& Decoder var. & 0.120171 & 0.538728 & 0.098943 \\
& Attenuation & 0.120057 & 0.530093 & 0.098822 \\
\bottomrule
\end{tabular}
\end{table}

\begin{table}[H]
\caption{Cross-architecture metric differences relative to the reference surrogate. Values are equal-trajectory-weight means with 95\% trajectory-bootstrap intervals. Bold rows exhibit a robust ranking inversion: lower field error and higher TKE error.}
\label{tab:architecture_conflict}
\centering
\small
\setlength{\tabcolsep}{4pt}
\begin{tabular}{llcc}
\toprule
Split & Model & $\Delta e_{\rel}$ & $\Delta e_{\tke}$ \\
\midrule
Selection & FNO & $-0.0384\,[-0.0493,-0.0277]$ & $+0.0067\,[-0.0169,0.0320]$ \\
& \textbf{CNO} & $\mathbf{-0.0189}\,[-0.0265,-0.0117]$ & $\mathbf{+0.1196}\,[0.1016,0.1369]$ \\
& Transolver & $+0.0351\,[0.0224,0.0482]$ & $+0.3067\,[0.2665,0.3401]$ \\
\midrule
Calibration & FNO & $-0.0385\,[-0.0495,-0.0279]$ & $+0.0071\,[-0.0154,0.0317]$ \\
& \textbf{CNO} & $\mathbf{-0.0191}\,[-0.0267,-0.0120]$ & $\mathbf{+0.1232}\,[0.1101,0.1366]$ \\
& Transolver & $+0.0350\,[0.0227,0.0475]$ & $+0.3033\,[0.2670,0.3313]$ \\
\bottomrule
\end{tabular}
\end{table}
For the CNO, 18 of 23 trajectory aggregates lie in the lower-field-error/higher-TKE-error quadrant on each subset. FNO provides a useful contrast: its field-error advantage is robust, whereas its small positive TKE-error difference has an interval spanning zero. This is why the main text treats CNO, not FNO, as the cross-architecture ranking inversion.

\section{Counterfactual and runtime provenance}
The counterfactual masks are constructed exclusively from the observed 20-frame PIV history; no future target enters mask selection. Masks are restricted to valid observed fluid cells and have equal area within each window, with ten deterministic random masks averaged for the random control. The realized 10\% energetic-mask fractions are 0.09973 on selection and 0.09979 on calibration.

Reference counterfactual and unperturbed predictions use the same frozen checkpoint/runtime, with window/target correspondence verified against frozen publication bundles. For the cross-model audit, FNO, CNO, and Transolver are loaded with the starting-kit normalization and output post-processing used for their real-finetuned checkpoints. Measured-channel cache parity was checked before perturbation. Mean absolute $u,v$ discrepancies were approximately $1.7\times10^{-5}$ for FNO, $2.0\times10^{-5}$ for CNO, and $1.4\times10^{-5}$ for Transolver, with maxima below $3.3\times10^{-4}$.

\section{Additional physical diagnostics and attenuation cross-fitting}
\label{app:physics}
Mean per-sample spatial TKE correlation changes from 0.81437 to 0.81427 on selection and from 0.81594 to 0.81594 on calibration after fixed attenuation. Thus the lower benchmark TKE error is not accompanied by improved spatial TKE alignment. Target-top-decile and pooled-error changes are reported in Sec.~3.3; Fig.~\ref{fig:attenuation} shows the corresponding signed-bias fields.

As an additional leakage audit, attenuation gains were also selected under five-fold trajectory-wise cross-fitting and applied only to held-out trajectories. The held-out energy mismatch robustly worsens, $\Delta|1-R_E|=+0.0390$ (95\% CI $[0.0218,0.0534]$), while the TKE-error change is $-0.00293$ ($[-0.00801,0.00154]$) and therefore not resolved away from zero. This negative robustness result supports the main-text interpretation of attenuation as a diagnostic stress test rather than a transferable physical correction.

\section{Computational environment and intervention provenance}
Counterfactual inference and publication analyses ran on a Dell PowerEdge R730 with Ubuntu 24.04.3 LTS, two Intel Xeon E5-2682 v4 CPUs (32 physical cores/64 threads), 251\,GiB RAM, and one NVIDIA A100-SXM4-40GB GPU. The environment used Python 3.10.20, PyTorch 2.2.2 with CUDA 12.1, NumPy 1.26.4, and cuDNN 8.9.2.

The decoder-variant checkpoint (internal identifier: v2A) was evaluated with its corresponding model definition under strict state-dictionary loading. Its role is limited to a within-lineage intervention; no general claim is made about horizon-conditioned decoders. The fixed attenuation gains $(g_u,g_v)=(0.975,0.935)$ were selected from a sweep that used both analysis subsets. The energy-ratio, signed-bias, spatial-correlation, pooled-TKE, and target-top-decile diagnostics were computed post hoc and were not objectives of that sweep.

\section{Interpretive boundaries and tests needed for stronger physical claims}\label{app:boundaries}
The input intervention removes temporal velocity fluctuations while holding each masked cell's observed temporal mean fixed. For a mask $\Omega$, the removed fluctuation-energy mass is
\begin{equation}
 E_{\mathrm{removed}}(\Omega)=\sum_{(x,y)\in\Omega} k_{\mathrm{hist}}(x,y).
\end{equation}
The equal-area energetic and random masks are not equal-energy controls: high-energy mask construction typically makes $E_{\mathrm{removed}}$ larger in the targeted intervention. Accordingly, the reported energetic-minus-random contrasts should be read as responses to different, physically defined input modifications, not as magnitude-adjusted feature importance. An energy-matched control or an analysis normalized by removed input energy would be needed to separate these explanations; neither result is reported here.

The masked histories are diagnostic inputs, not solutions of the governing equations or physically verified flow interventions. Their interfaces introduce spatial discontinuities that can interact with spectral operators. Smooth transition masks, solver-generated counterfactuals, and comparison with actual experimental consequences would be required before asserting physically faithful counterfactual dynamics. This study has no counterfactual future ground truth and makes no such assertion.

Finally, velocity-field error, two-component resolved fluctuation-energy error, spatial TKE correlation, regional TKE MSE, and domain-summed fluctuation-energy ratio each measure different aspects of forecast agreement with available PIV observations. A disagreement between these quantities is evidence that a scalar benchmark score provides an incomplete account of the measured flow statistics; it does not by itself determine which model would perform better for a downstream physical application. The four surrogates, shared trajectory support, and attenuation tuned on both subsets constrain wider generalization. Energy-matched and smooth-mask tests, solver-validated interventions, additional physical systems and trajectory-disjoint evaluation, and other training families (including generative approaches) are promising next steps. See \citep{oommen2026turbulence} for an example of generative turbulence modelling.

\end{document}